\pdfoutput=1

\documentclass[11pt]{article}

\usepackage{acl}

\usepackage{times}
\usepackage{latexsym}

\usepackage[T1]{fontenc}

\usepackage[utf8]{inputenc}

\usepackage{microtype}

\usepackage{graphicx}
\usepackage{paralist}
\usepackage{multirow}
\usepackage[normalem]{ulem}
\useunder{\uline}{\ul}{}

\title{Hate Speech and Offensive Language Detection in Bengali}

\author{Mithun Das,  Somnath Banerjee,  Punyajoy Saha,  Animesh Mukherjee \\
  Indian Institute of Technology Kharagpur, West Bengal, India \\
  \texttt{mithundas@iitkgp.ac.in, som.iitkgpcse@kgpian.iitkgp.ac.in,} \\ \texttt{punyajoys@iitkgp.ac.in,  animeshm@cse.iitkgp.ac.in} \\}

\begin{document}
\maketitle
\begin{abstract}
Social media often serves as a breeding ground for various hateful and offensive content. Identifying such content on social media is crucial due to its impact on the race, gender, or religion in an unprejudiced society. However, while there is extensive research in hate speech detection in English, there is a gap in hateful content detection in low-resource languages like Bengali. Besides, a current trend on social media is the use of Romanized Bengali for regular interactions. To overcome the existing research's limitations, in this study, we develop an annotated dataset of 10K Bengali posts consisting of 5K actual and 5K Romanized Bengali tweets. We implement several baseline models for the classification of such hateful posts. We further explore the interlingual transfer mechanism to boost classification performance. Finally, we perform an in-depth error analysis by looking into the misclassified posts by the models. While training actual and Romanized datasets separately, we observe that XLM-Roberta performs the best. Further, we witness that on joint training and few-shot training, MuRIL outperforms other models by interpreting the semantic expressions better. We make our code and dataset public for others\footnote{\url{https://github.com/hate-alert/Bengali_Hate}}.
\end{abstract}

\section{Introduction}

\noindent Social media websites like Twitter and Facebook have brought billions of people together and given them the opportunity to share their thoughts and opinions rapidly. On the one hand, it has facilitated communication and the growth of social networks; on the other, it has been exploited to propagate misinformation, violence, and hate speech~\cite{hate-speech-websci-19,das2020hate} against users based on their gender, race, religion, or other characteristics. If such content is left unaddressed, it may result in widespread conflict and violence, raising concerns about the safety of human rights, the rule of law, and freedom of speech, all of which are crucial for the growth of an unprejudiced democratic society~\cite{rizwan2020hate}. Organizations such as Facebook have been blamed for being a forum for instigating anti-Muslim violence in Sri Lanka that resulted in the deaths of three individuals\footnote{\url{https://tinyurl.com/sriLankaRiots}}, and a UN report accused them of disseminating hate speech in a way that contributed significantly to the plausible genocide of the Rohingya population in Myanmar\footnote{\url{https://www.reuters.com/investigates/special-report/myanmar-facebook-hate}}.

In order to reduce the dissemination of such harmful content, these platforms have developed certain guidelines\footnote{\label{twitter_violation}\url{https://help.twitter.com/en/rules-and-policies/hateful-conduct-policy}}  that the users of these platforms ought to comply with. If these rules aren't followed, the post can get deleted, or the user's account might get suspended. Even to diminish the harmful content from their forum, these platforms engage moderators~\cite{newton_2019} to manually review the posts and preserve the platform as wholesome and people-friendly. However, this moderation strategy is confined by the moderators' speed, jargon, capability to understand the development of slang, and familiarity with multilingual content. Moreover, due to the sheer magnitude of data streaming, it is also an ambitious endeavor to examine each post manually and filter out such harmful content. Hence, an automated technique for detecting hate speech and offensive language is extremely necessary and inevitable.

It has already been witnessed that Facebook vigorously eliminated a considerable amount of malicious content from its platforms even before users reported it~\cite{robertson_2020}. However, the hindrance is that these platforms can detect harmful content in certain popular languages such as English, Spanish, etc.~\cite{perrigo_2019} So far, several investigations have been conducted to identify hate speech automatically, focusing mainly on the English language; therefore, an effort is required to determine and diminish such hateful content in low-resource languages.

With more than 210 million speakers, Bengali is the seventh most widely spoken language\footnote{\url{https://www.berlitz.com/en-uy/blog/most-spoken-languages-world}}, with around 100 million Bengali speakers in Bangladesh and 85 million in India. Apart from Bangladesh and India, Bengali is spoken in many countries, including the United Kingdom, the United States, and the Middle East\footnote{\url{https://www.britannica.com/topic/Bengali-language}}. Also, a current trend on social media platforms is that apart from actual Bengali, people tend to write Bengali using Latin scripts(English characters) and often use English phrases in the same conversation. This unique and informal communication dialect is called code-mixed Bengali or Roman Bengali. Code-mixing makes it easier for speakers to communicate with one another by providing a more comprehensive range of idioms and phrases. However, as emphasized by Chittaranjan et al.~\cite{chittaranjan2014word}, this has made the task of creating NLP tools more challenging. Along with these challenges, the challenges specific to identifying hate speech in Roman Bengali contain the following: \textit{Absence of a hate speech dataset}, \textit{Lack of benchmark models}. Thus, there is a need to develop open efficient datasets and models to detect hate speech in Bengali. Although few studies have been conducted in developing Bengali hate speech datasets, most of these have been crawled with comments from Facebook pages, and all of them are in actual Bengali. Hence, there is a need for developing more benchmarking datasets considering other popular platforms. To address these limitations, in this study, we make the following contributions.

\begin{compactenum}
    \item [-] First, we create a gold-standard dataset of 10K tweets among which 5K tweets are actual Bengali and 5K tweets are Roman Bengali.
    \item [-] Second, we implement several baseline models to identify such hateful and offensive content automatically for both actual \& Roman Bengali tweets.
    \item [-] Third, we explore several interlingual transfer mechanisms to boost the classification performance. 
    \item  [-] Finally, we perform in-depth error analysis by looking into a sample of posts where the models mis-classify some of the test instances.
\end{compactenum}

\section{Related Work}

\label{sec:related_works}

Over the past few years, research around automated hate speech detection has been evolved tremendously. The earlier effort in developing resources for the hate speech detection was mainly focused around English language~\cite{waseem2016hateful,davidson2017automated,founta2018large}. Recently, in an effort to create multilingual hate speech datasets, several shared task competitions have been organized (HASOC~\cite{mandl2019overview}, OffensEval~\cite{zampieri2019semeval},, TRAC~\cite{kumar2020evaluating}, etc.), and multiple datasets such as Hindi~\cite{modha2021overview}, Danish~\cite{sigurbergsson2020offensive}, Greek~\cite{pitenis2020offensive}, Turkish~\cite{ccoltekin2020corpus}, Mexican Spanish~\cite{aragon2019overview}, etc. have been made public. There is also some work to detect hate speech in actual Bengali.  Ismam et al. \cite{ishmam2019hateful} collected and annotated 5K comments from Facebook into six classes-\textit{inciteful}, \textit{hate speech}, \textit{religious hatred}, \textit{communal attack}, \textit{religious comments}, and \textit{political comments}. However,the dataset is not publicly available. Karim et al.~\cite{karim2021deephateexplainer} provided a dataset of 8K hateful posts collected from multiple sources such as Facebook, news articles, blogs, etc. One of the problems with this dataset is that all comments are part of any hate class(\textit{personal}, \textit{geopolitical}, \textit{religious}, and \textit{political}), so we cannot build hate speech detection models using this dataset to screen out hate speech. Romim et al.~\shortcite{romim2021hate} curated a dataset of 30K comments, making it one of the most extensive datasets for hateful statements. The author achieved 87.5\% accuracy on their test dataset using the SVM model.  However, these datasets do not consider Roman Bengali posts, a prevalent communication method on social media nowadays.

With regards to the detection systems, earlier methods examined simple linguistic features such as character and word n-grams, POS tags, tf-idf with a traditional classifier such as  LR, SVM, Decision Tree, etc~\cite{davidson2017automated}. With the development of larger datasets, researchers have shifted to data-hungry complex models such as deep learning~\cite{Pitsilis2018DetectingOL,zhang2018detecting} and graph embedding techniques to enrich the classifier performance. 

Recently, transformer-based~\cite{Vaswani2017AttentionIA} language models such as BERT, XLM-RoBERTa~\cite{Devlin2019BERTPO} are becoming quite popular in several downstream tasks. It has already been observed that these transformer-based models outperform several earlier deep learning models ~\cite{mathew2020hatexplain}. Having observed these transformer-based models' superior performance, we focus on building these models for our classification task. 

Further, researchers have begun to explore few shot classifications. One of the most popular techniques for few-shot classification is transfer learning\ - where a model (pre-trained in a similar domain) is further fine-tuned on a few labeled samples in the target domain~\cite{alyafeai2020survey}. Keeping these experiments in mind, we also examine the ability of transfer learning capabilities between actual and Roman Bengali data.

\begin{table}[t]
\centering
\begin{tabular}{l|l|l|l|}
\cline{2-4}
 & \textbf{Actual} & \textbf{Roman} & \textbf{Total} \\ \hline
\multicolumn{1}{|l|}{\textbf{Hateful}} & 825 & 510 & 1,335 \\ \hline
\multicolumn{1}{|l|}{\textbf{Offensive}} & 1,341 & 2,063 & 3,404 \\ \hline
\multicolumn{1}{|l|}{\textbf{Normal}} & 2,905 & 2,534 & 5,439 \\ \hline
\multicolumn{1}{|l|}{\textbf{Total}} & 5,071 & 5,107 & 10,178 \\ \hline
\end{tabular}
\caption{Dataset Statistics of both Actual and Roman tweets.}
\label{dataStat}
\end{table}

\section{Dataset Creation}
\label{sec:dataset}

In this section, we provide the data collection procedure, annotation strategies we have followed and the statistics of the collected dataset.

\subsection{Dataset collection and sampling}

In this paper, we collect our dataset from \textbf{Twitter}. Despite Hatebase.org maintaining the most extensive collection of multilingual hateful words, it still lacks such lexicon base for Bengali\footnote{\url{https://hatebase.org/}}. To sample Bengali (actual and romanized) tweets for annotation, we create a lexicon of 74 abusive terms\footnote{\label{footnoteBengali}\url{https://tinyurl.com/bengaliHate}}). These lexicons consist of derogatory keywords/slurs targeting individuals or different protected communities. We also include words based on the name of the targeted communities. The choice to add names of targeted communities is made in order to extract random hateful/offensive tweets that do not contain any abusive words. Using Twitter API, we searched for tweets containing phrases from the lexicons, which resulted in a sample of 500K tweets for actual Bengali and 150K tweets for Roman Bengali. To evade problems related to user distribution bias, as highlighted by Arango et al.~\cite{arango2019hate}, we limit a maximum of 75 tweets per user. We also do not use more than 500 tweets per month to avoid event-specific tweets in our dataset.

\subsection{Annotation  procedure} 
We employed four undergraduate students for our annotation task. All undergraduate students are Computer Science majors and native Bengali speakers. They have been recruited voluntarily through departmental emails and compensated via an Amazon gift card. Two Ph.D. students led the annotation process as expert annotators. Both expert annotators had previous experience working with malicious content on social media. Each tweet in our dataset contains two kinds of annotations: first whether the text is hate speech, offensive speech, or normal; second, the target communities in the text. This additional annotation of the target community can help us measure bias in the model. Table \ref{tab:targetDetails} lists the target groups we have considered.

\noindent{\textbf{Annotation guidelines:}} The annotation scheme stated below constitute the main guidelines for the annotators, while a codebook ensured common understanding of the label descriptions. We construct our codebook (which consists the annotation guidelines for identifying hateful and offensive tweets based on the definitions summarized as follows.

\begin{compactenum}
    \item [-] \textbf{Hate speech:} \textit{Hate speech is a language used to express hatred toward a targeted individual or group or is intended to be derogatory, humiliating, or insulting to the group members based on attributes such as race, religion, ethnic origin, sexual orientation, disability, caste, geographic location or gender.}

    \item [-] \textbf{Offensive:}\textit{ Offensive speech uses profanity, strongly impolite, rude, or vulgar language expressed with fighting or hurtful words to insult a targeted individual or group.}

    \item [-] \textbf{Normal:} \textit{This contains tweets that do not fall into the above categories.}
\end{compactenum}

\begin{table*}
\centering
  \includegraphics[width=\linewidth]{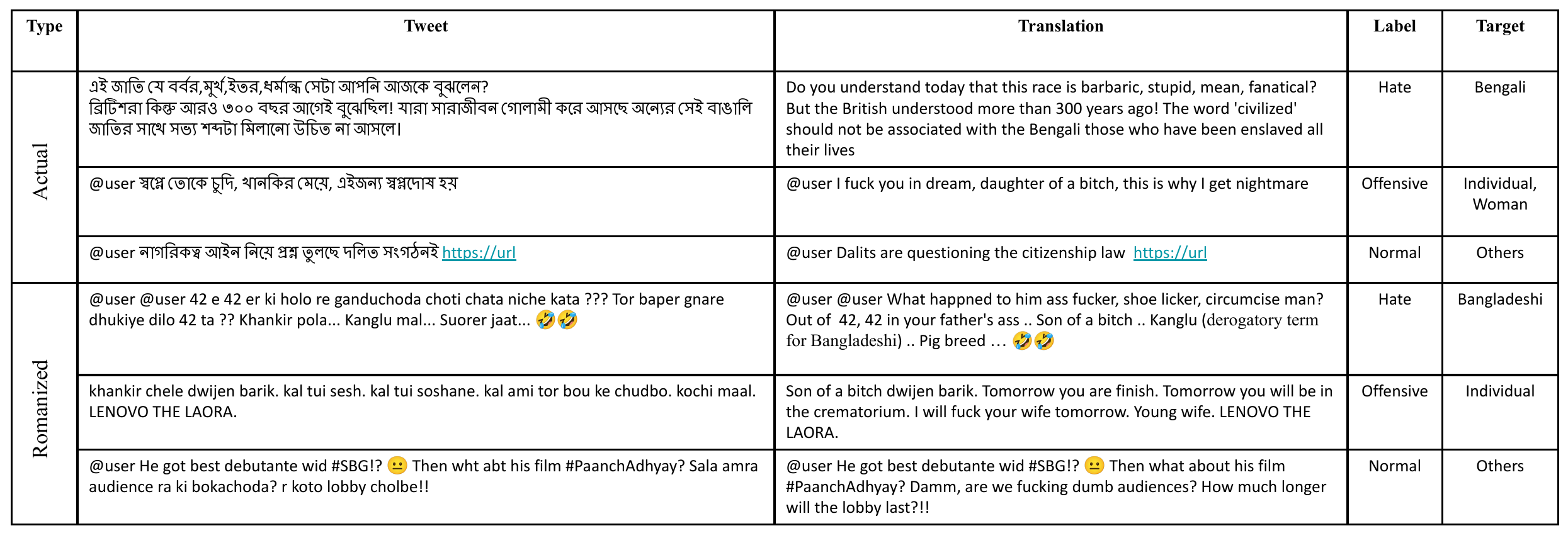}
    \caption{Samples of Actual and Roman Bengali tweets for each label from the dataset}
  \label{tbl:bnHate}
\end{table*}

\begin{table}
\centering
\resizebox{\linewidth}{!}{%
\begin{tabular}{|l|l|}
\hline
\textbf{Target Groups} & \textbf{Categories}                                                                       \\ \hline
Gender                 & Men, Women, Trans.                                                                         \\
Linguistic Community   & Bengali, Bihari.                                                                           \\
National Origin        & Indian, Bangladeshi, Pakistani.                                                            \\
Religion               & Hindu, Islam.                                                                             \\
Miscellaneous          & \begin{tabular}[c]{@{}l@{}}Individual, Political, Disabled, \\ Dalit, Others.\end{tabular} \\ \hline
\end{tabular}%
}
\caption{Target groups considered for the annotation.}
\label{tab:targetDetails}
\end{table}

\subsection{Dataset creation steps}

As a first step for creating the dataset, we required a pilot gold-label dataset to instruct the annotators. Initially, the expert annotators annotated 100 tweets, out of which 30 were hateful, 35 were offensive, and the rest 35 tweets were normal.

\noindent{\textbf{Pilot annotation:}} Each annotator was given 30 tweets from the gold-label dataset in the pilot task. They were asked to classify hate/offensive speech and identify the target community (if any). They were provided the annotation codebook with multiple examples for the labeling process to understand the task clearly. They were asked to keep the annotation guidelines open while doing the annotation to have better clarity about the labeling scheme. After the annotators finished this set, we consulted the incorrect annotations in their set with them. This activity further trained the annotators and helped to fine-tune the annotation scheme. In addition, we collected feedback from annotators to enrich the main annotation task.

\noindent{\textbf{Main annotation:}} After the training process, we proceeded with the main annotation task. For this task, we use the open-source platform Docanno\footnote{\url{https://github.com/doccano/doccano}}, deployed on a Heroku instance. We provided a secure account to each annotator where they could annotate and track their progress. Two independent annotators annotated each tweet. Based on the guidelines, they were instructed to read the entire tweet and select the appropriate category (hate speech, offensive, or standard). Initially, we started with a small batch of 100 tweets and later expanded it to 500 tweets as the annotations became more efficient. We tried to preserve the annotators' agreement by pointing out some errors they made in the previous batch. Since hate/offensive speech is highly polarizing and adverse, the annotators were given plenty of time to complete the annotations. On completion of each set of annotations, if there was a mismatch between two annotators, one of the expert annotators annotated the same tweet to break the tie. For the cases where all the three annotators chose a separate class, we did not consider these tweets for further analysis. To determine the target community of a tweet, we combine the annotated targets.

Exposure to online abuse could lead to unhealthy mental health issues\footnote{\url{https://www.theguardian.com/technology/2017/jan/11/microsoft-employees-child-abuse-lawsuit-ptsd}}\cite{ybarra2006examining}. Therefore, the annotators were recommended to take periodic breaks and not do the annotations in one sitting. Besides, we also had weekly meetings with them to ensure the annotations did not have any effect on their mental health.

\noindent{\textbf{Final dataset:}} Table \ref{dataStat} notes our final dataset statistics. It consists of 5,071 actual Bengali tweets (out of which 825 have been labelled as hateful, 1,341 are offensive, and 2,905 tweets are normal) and 5,107
Roman Bengali tweets (out of which 510 tweets are hateful, 2,063 tweets are offensive, and 2,534 tweets are normal). We achieved an inter-annotator agreement of 0.696 using Krippendorff's $\alpha$ which is better than the agreement score on other related hate speech tasks~\cite{ousidhoum2019multilingual,guest2021expert}. In Table~\ref{tbl:bnHate} we have shown some examples of Bengali hate speech and offensive language that we have annotated.

\section{Methodology}
\subsection{Baseline models}
In this section, we discuss the models we implement for automatic detection of hate speech. We experimented with a wide range of models for our use case.

\noindent\textbf{m-BERT}~\cite{Devlin2019BERTPO} is a stack of transformer encoder layers consisting of 12 ``attention heads" with self-attention mechanisms. It is pre-trained on 104 languages using a masked language modeling (MLM) objective with the crawled Wikipedia data. To fine-tune m-BERT, we include a fully connected layer with the output corresponding to the CLS token in the input. Typically, the expression of the sentence provided to the model is retained in this CLS token output. In hate speech, the m-BERT model has been well studied, outperformed several baselines, and is considered state-of-the-art.

\noindent\textbf{XLM-Roberta}~\cite{conneau2019unsupervised} is another form of Transformer model, pre-trained on 2.5TB of filtered CommonCrawl data containing 100 languages. XLM-R was trained using a lot more data compared to m-BERT. Similar to BERT, it is a stack of transformer encoder layers with 12 ``attention heads" and can handle at max 512 tokens. 

\noindent\textbf{IndicBERT}~\cite{kakwani2020indicnlpsuite} is a multilingual ALBERT model~\cite{lan2019albert} (a recent derivative of BERT) trained on large-scale corpora, covering 12 major Indian languages. It is pre-trained on 9 billion tokens and evaluated on a set of diverse tasks. Unlike m-BERT, XLM-Roberta, IndicBERT has around 10x fewer parameters and still manages to deliver state-of-the-art performance on several tasks.

\noindent\textbf{MuRIL}~\cite{khanuja2021MuRIL} stands for Multilingual Representations for Indian Languages and aims to enrich reciprocity from one language to another. This model uses a BERT base architecture pre-trained from scratch using the Common Crawl, Wikipedia, PMINDIA, and Dakshina corpora for 17 Indian languages and their transliterated counterparts.

\subsection{Interlingual transfer mechanisms}

One of the main attractions of transformer-based models is their potential to strengthen model transfer via several mechanisms. This can be especially beneficial for enhancing learning performance in low-resource languages like Bengali. In order to evaluate the extent to which language similarity improves transfer learning performance, we perform the following tests.\footnote{Although the discussed models have been pre-trained using multiple languages, fine-tuning has been done using the Bengali language dataset.}

\noindent{\textbf{ELFI (Each language for itself):}} In this situation, we use the same language's data for training, validation, and testing. This scenario typically appears in the real world, where monolingual datasets are frequently utilized to build classifiers for a particular language. Despite the anticipated high labeling costs, this gives an idea of the most achievable classification performance.

\noindent{\textbf{Joint training:}} In this setting, we integrate both actual \& Roman Bengali posts to train all the transformer-based models. The notion is that even though the characters used to represent both languages are different, their semantic content is mostly the same. Hence, it gives an idea of whether jointly training the models can benefit learning the better semantic representation of a particular post for determining the corresponding label of the post.

\noindent{\textbf{Model transfer:}} In this scenario, the models are trained with one language (source language) and evaluated in another language (target language). In the zero-shot setting, no instances from the target language have been used while training (\textbf{MTx0}). In a related few-shot setting, we allow $n$ = 32, 64, and 128 posts per label from the available gold target instances to fine-tune the existing models (trained in another language). These are named \textbf{MTx32, MTx64} and \textbf{MTx128}.

\noindent{\textbf{Language transfer:}} In this setting, we translate the Bengali posts to English using Google Translate tool\footnote{\label{footnotetranslator}\url{https://cloud.google.com/translate}} and do the entire training, testing on the translated instances. We do this to check if language space has been transformed for a task, how model's performance varies. 

\noindent{\textbf{Joint training with language transfer:}} In this scenario, we combine the translated Bengali and Roman Bengali posts, to train all the transformer based models. The motivation behind this experiment is that, in case of romanized Bengali data, people use English words/sentences in their posts for ease of writing. Thus, we perform this experiment to determine whether adding translated Bengali data points will further improve the performance of the classification or not.

\subsection{Experimental setup}
All the models are evaluated using the same 70:10:20 train, validation, and test split, stratified by class across the splits. For the model transfer evaluation, we use 32, 64, and 128 training data points from each class to train the model in another language. We create three such different random sets for the target dataset to have a more robust assessment and report the average performance. The models were run for 10 epoch with Adam optimizer, batch\_size = 16, learning\_rate = $2e-5$ and adam\_epsilon = $1e-8$. In addition, we set the number of tokens $n$ = 400 for all the models.

\subsection{Evaluation metric}

To remain consistent with existing literature, we evaluate our models in terms of \textbf{accuracy}, \textbf{F1-score} and \textbf{AUROC} score. These metrics together should be able to thoroughly evaluate the classification performance in distinguishing among the three classes, e.g., hate, offensive and normal. For zero-shot and few-shot settings, we report only \textbf{macro F1-score} due to paucity of space. We also highlight the best performance using \textbf{bold} and second best using \underline{underline}.

\begin{table*}
\centering
\scriptsize
\begin{tabular}{l|ccccc|c|ccccc|}
\cline{2-6} \cline{8-12}
                                         & \multicolumn{5}{c|}{\textbf{Actual Bengali}}                                                                                                                            &                   & \multicolumn{5}{c|}{\textbf{Roman Bengali}}                                                                                                                             \\ \hline
\multicolumn{1}{|l|}{\textbf{Model}}     & \multicolumn{1}{c|}{\textbf{Acc}}   & \multicolumn{1}{c|}{\textbf{M-F1}}  & \multicolumn{1}{c|}{\textbf{F1(H)}} & \multicolumn{1}{c|}{\textbf{F1(O)}} & \textbf{AUROC} & \multirow{5}{*}{} & \multicolumn{1}{c|}{\textbf{Acc}}   & \multicolumn{1}{c|}{\textbf{M-F1}}  & \multicolumn{1}{c|}{\textbf{F1(H)}} & \multicolumn{1}{c|}{\textbf{F1(O)}} & \textbf{AUROC} \\ \cline{1-6} \cline{8-12} 
\multicolumn{1}{|l|}{\textbf{m-BERT}}    & \multicolumn{1}{c|}{0.813}          & \multicolumn{1}{c|}{0.795}          & \multicolumn{1}{c|}{\textbf{0.824}}  & \multicolumn{1}{c|}{0.693}          & {\ul 0.917}    &                   & \multicolumn{1}{c|}{0.840}          & \multicolumn{1}{c|}{0.789}          & \multicolumn{1}{c|}{{\ul 0.658}}     & \multicolumn{1}{c|}{0.840}          & {\ul 0.910}    \\ \cline{1-6} \cline{8-12} 
\multicolumn{1}{|l|}{\textbf{XLM}}       & \multicolumn{1}{c|}{\textbf{0.830}} & \multicolumn{1}{c|}{\textbf{0.803}} & \multicolumn{1}{c|}{0.812}           & \multicolumn{1}{c|}{\textbf{0.717}} & \textbf{0.919} &                   & \multicolumn{1}{c|}{\textbf{0.858}} & \multicolumn{1}{c|}{\textbf{0.805}} & \multicolumn{1}{c|}{\textbf{0.666}}  & \multicolumn{1}{c|}{\textbf{0.857}} & \textbf{0.924} \\ \cline{1-6} \cline{8-12} 
\multicolumn{1}{|l|}{\textbf{MuRIL}}     & \multicolumn{1}{c|}{{\ul 0.817}}    & \multicolumn{1}{c|}{{\ul 0.797}}    & \multicolumn{1}{c|}{{\ul 0.816}}     & \multicolumn{1}{c|}{{\ul 0.704}}    & 0.887          &                   & \multicolumn{1}{c|}{0.843}          & \multicolumn{1}{c|}{0.788}          & \multicolumn{1}{c|}{0.646}           & \multicolumn{1}{c|}{0.841}          & 0.897          \\ \cline{1-6} \cline{8-12} 
\multicolumn{1}{|l|}{\textbf{IndicBERT}} & \multicolumn{1}{c|}{0.790}          & \multicolumn{1}{c|}{0.767}          & \multicolumn{1}{c|}{0.788}           & \multicolumn{1}{c|}{0.656}          & 0.896          &                   & \multicolumn{1}{c|}{{\ul 0.846}}    & \multicolumn{1}{c|}{{\ul 0.793}}    & \multicolumn{1}{c|}{0.651}           & \multicolumn{1}{c|}{{\ul 0.846}}    & 0.908          \\ \hline
\end{tabular}
\caption{Performance on Both Actual \& Roman Bengali Datasets. XLM:XLM-Roberta, M: Macro, Acc: Accuracy, H: Hate, O: Offensive.}
\label{bengaliCodeStandalone}
\end{table*}

\begin{table*}
\centering
\scriptsize
\begin{tabular}{l|ccccc|c|ccccc|}
\cline{2-6} \cline{8-12}
 & \multicolumn{5}{c|}{\textbf{Actual Bengali}} & \multicolumn{1}{l|}{} & \multicolumn{5}{c|}{\textbf{Roman Bengali}} \\ \hline
\multicolumn{1}{|l|}{\textbf{Model}} & \multicolumn{1}{c|}{\textbf{Acc}} & \multicolumn{1}{c|}{\textbf{M-F1}} & \multicolumn{1}{c|}{\textbf{F1(H)}} & \multicolumn{1}{c|}{\textbf{F1(O)}} & \textbf{AUROC} & \multirow{5}{*}{} & \multicolumn{1}{c|}{\textbf{Acc}} & \multicolumn{1}{c|}{\textbf{M-F1}} & \multicolumn{1}{c|}{\textbf{F1(H)}} & \multicolumn{1}{c|}{\textbf{F1(O)}} & \textbf{AUROC} \\ \cline{1-6} \cline{8-12} 
\multicolumn{1}{|l|}{\textbf{m-BERT}} & \multicolumn{1}{c|}{{\ul 0.829}} & \multicolumn{1}{c|}{{\ul 0.800}} & \multicolumn{1}{c|}{{\ul 0.831}} & \multicolumn{1}{c|}{0.684} & \textbf{0.914} &  & \multicolumn{1}{c|}{0.845} & \multicolumn{1}{c|}{0.789} & \multicolumn{1}{c|}{0.647} & \multicolumn{1}{c|}{0.830} & \textbf{0.928} \\ \cline{1-6} \cline{8-12} 
\multicolumn{1}{|l|}{\textbf{XLM}} & \multicolumn{1}{c|}{0.819} & \multicolumn{1}{c|}{0.794} & \multicolumn{1}{c|}{0.805} & \multicolumn{1}{c|}{{\ul 0.701}} & {\ul 0.912} &  & \multicolumn{1}{c|}{\textbf{0.865}} & \multicolumn{1}{c|}{\textbf{0.810}} & \multicolumn{1}{c|}{{\ul 0.666}} & \multicolumn{1}{c|}{\textbf{0.867}} & {\ul 0.918} \\ \cline{1-6} \cline{8-12} 
\multicolumn{1}{|l|}{\textbf{MuRIL}} & \multicolumn{1}{c|}{\textbf{0.833}} & \multicolumn{1}{c|}{\textbf{0.808}} & \multicolumn{1}{c|}{\textbf{0.835}} & \multicolumn{1}{c|}{\textbf{0.704}} & 0.895 &  & \multicolumn{1}{c|}{{\ul 0.850}} & \multicolumn{1}{c|}{{\ul 0.800}} & \multicolumn{1}{c|}{\textbf{0.670}} & \multicolumn{1}{c|}{{\ul 0.842}} & 0.904 \\ \cline{1-6} \cline{8-12} 
\multicolumn{1}{|l|}{\textbf{IndicBERT}} & \multicolumn{1}{c|}{0.785} & \multicolumn{1}{c|}{0.769} & \multicolumn{1}{c|}{0.807} & \multicolumn{1}{c|}{0.658} & 0.900 &  & \multicolumn{1}{c|}{0.817} & \multicolumn{1}{c|}{0.767} & \multicolumn{1}{c|}{0.637} & \multicolumn{1}{c|}{0.808} & 0.890 \\ \hline
\end{tabular}
\caption{Performance of Both Actual \& Roman Bengali Datasets on Joint Training. XLM:XLM-Roberta, M: Macro, Acc: Accuracy, H: Hate, O: Offensive.}
\label{jointTraining}
\end{table*}

\section{Results}
In this section, we discuss the findings of our experiments.

\subsection{Performnace of ELFI}
In Table \ref{bengaliCodeStandalone}, we report the performance of all the models for actual \& Roman Bengali. We observe for both of these, XLM-Roberta performs the best in terms of accuracy and macro-F1 score. Followed by XLM-Roberta, MuRIL performed the second best for the actual Bengali. For the hate class m-BERT does slightly better than XLM-Roberta in terms of F1-score. For Roman Bengali, IndicBERT performs next to XLM-Roberta.

\subsection{Performance of joint training}
Here we investigate the importance of joint training. Even though both the actual \& Roman Bengali is written using different characters, semantic expression of both the languages are same. Table \ref{jointTraining} summarizes the performance of different models when trained jointly. We observe some improvements in the joint training models. In particular, MuRIL, which is pretrained on both Indian languages and their transliterated counterparts, is able to interrelate the semantics of the actual \& Roman Bengali sentences. We notice that for actual Bengali, MuRIL performs the best with Macro-F1 score of 0.808 (and accuracy of 0.833), followed by m-BERT with Macro-F1 score of 0.800 (and accuracy of 0.829). For the Roman Bengali though, XLM-Roberta still performs the best (with Macro F1-score of 0.810), MuRIL performs very close to it and in fact better for the hate class F1-score.

\subsection{Performance of model transfer}
In this scenario, we investigate the power of existing fine-tuned models. The idea is to understand how these models are generalized across the same language, having same semantic content, but are written using different characters/words. We report our results in Table \ref{zeroFewShotTraining}. 

In \textbf{zero-shot} setting we observe, when the model is trained on actual Bengali and tested on Roman Bengali, m-BERT performs the best (with macro F1 score of 0.390) among all the models followed by IndicBERT (Macro F1 score 0.319). On the other hand, when trained on Roman Bengali and tested on actual Bengali, MuRIL performs the best (macro F1 score 0.414) among all the models followed by IndicBERT (macro F1 score 0.397). An interesting thing to note is that although the XLM-Roberta performs best in monolingual settings, it is not performing well in the model transfer setup.

\begin{table}
\centering
\scriptsize
\begin{tabular}{|ccccc|}
\hline
\multicolumn{5}{|c|}{\textbf{Actual Bengali Model's Performance on Roman Bengali}}                                                                                                                                                                                                                                                                                                                   \\ \hline
\multicolumn{1}{|c|}{\textbf{Model}}     & \multicolumn{1}{c|}{\textbf{\begin{tabular}[c]{@{}c@{}}Zero-Shot\\ (MTx0)\end{tabular}}} & \multicolumn{1}{c|}{\textbf{\begin{tabular}[c]{@{}c@{}}Few-Shot\\ (MTx32)\end{tabular}}} & \multicolumn{1}{c|}{\textbf{\begin{tabular}[c]{@{}c@{}}Few-Shot\\ (MTx64)\end{tabular}}} & \textbf{\begin{tabular}[c]{@{}c@{}}Few-Shot\\ (MTx128)\end{tabular}} \\ \hline
\multicolumn{1}{|c|}{\textbf{m-BERT}}     & \multicolumn{1}{c|}{\textbf{0.390}}    & \multicolumn{1}{c|}{\textbf{0.530}}                                                               & \multicolumn{1}{c|}{\underline{0.655}}                                                               & \underline{0.692}                                                                \\ \hline
\multicolumn{1}{|c|}{\textbf{XLM-Roberta}} & \multicolumn{1}{c|}{0.230}                                                               & \multicolumn{1}{c|}{0.456}                                                               & \multicolumn{1}{c|}{0.570}                                                               & 0.668                                                                \\ \hline
\multicolumn{1}{|c|}{\textbf{MuRIL}}     & \multicolumn{1}{c|}{0.269}                                                               & \multicolumn{1}{c|}{\underline{0.507}}                                                               & \multicolumn{1}{c|}{\textbf{0.671}}                                                               & \textbf{0.751}                                                                \\ \hline
\multicolumn{1}{|c|}{\textbf{IndicBERT}} & \multicolumn{1}{c|}{\underline{0.319}}                                                               & \multicolumn{1}{c|}{0.332}                                                               & \multicolumn{1}{c|}{0.355}                                                               & 0.462                                                                \\ \hline
\multicolumn{5}{|c|}{\textbf{Roman Bengali Model's Performance on Actual Bengali}}                                                                                                                                                                                                                                                                                                                           \\ \hline
\multicolumn{1}{|c|}{\textbf{Model}}     & \multicolumn{1}{c|}{\textbf{\begin{tabular}[c]{@{}c@{}}Zero-Shot\\ (MTx0)\end{tabular}}} & \multicolumn{1}{c|}{\textbf{\begin{tabular}[c]{@{}c@{}}Few-Shot\\ (MTx32)\end{tabular}}} & \multicolumn{1}{c|}{\textbf{\begin{tabular}[c]{@{}c@{}}Few-Shot\\ (MTx64)\end{tabular}}} & \textbf{\begin{tabular}[c]{@{}c@{}}Few-Shot\\ (MTx128)\end{tabular}} \\ \hline
\multicolumn{1}{|c|}{\textbf{m-BERT}}     & \multicolumn{1}{c|}{0.268}                                                               & \multicolumn{1}{c|}{0.449}                                                               & \multicolumn{1}{c|}{0.608}                                                               & \underline{0.691}                                                                \\ \hline
\multicolumn{1}{|c|}{\textbf{XLM-Roberta}} & \multicolumn{1}{c|}{0.299}                                                               & \multicolumn{1}{c|}{\underline{0.542}}                                                               & \multicolumn{1}{c|}{\underline{0.613}}                                                               & 0.664                                                                \\ \hline
\multicolumn{1}{|c|}{\textbf{MuRIL}}     & \multicolumn{1}{c|}{\textbf{0.414}}                                                               & \multicolumn{1}{c|}{\textbf{0.575}}                                                               & \multicolumn{1}{c|}{\textbf{0.645}}                                                               & \textbf{0.709}                                                                \\ \hline
\multicolumn{1}{|c|}{\textbf{IndicBERT}} & \multicolumn{1}{c|}{\underline{0.397}}                                                               & \multicolumn{1}{c|}{0.463}                                                               & \multicolumn{1}{c|}{0.508}                                                               & 0.557                                                                \\ \hline
\end{tabular}
\caption{Performance of Zero-shot \& Few-shot Learning.}
\label{zeroFewShotTraining}
\end{table}

To further investigate, how the performance of these models would vary, we conduct a second stage of fine-tuning. In this setting we use the existing trained model in actual Bengali and further fine-tune it with $n$ samples of Roman Bengali data points per label (and vice-versa). we repeat the subset sampled data selection with 3 different random sets and report the average performance. This will help to reduce performance variations across different sets. In general We observed with the increasing data points the performance of all models has improved. 
\begin{compactenum}
    \item [-] \textbf{Actual} $\rightarrow$ \textbf{Roman}: We observe further fine-tuning the model with 32 instances, m-BERT performs the best followed by MuRIL. While increasing these instances, MuRIL outperforms all other models. Only with 128 instances per label, MuRIL achieves macro F1-Score of 0.751.

    \item [-] \textbf{Roman} $\rightarrow$ \textbf{Actual}: We see MuRIL outperforms all other models. Followed by MuRIL, XLM-Robera performed the second best for 32 and 64 instances and for 128 instances m-BERT is the second best.
\end{compactenum}

\subsubsection{Performance of language transfer}
Here we investigate the importance of gold instances\footnote{raw labeled data} in a low resource language. We do so by transforming the language space. We translate the Bengali datasets to English and do training, testing on the translated dataset. In Table \ref{bengaliEnglisgTrans} we report the results of all the models. Although XLM-Roberta outperforms all other models, an important point to note is that its performance (Macro-F1 score 0.764) is much lower compared to the model trained on the gold (i.e., actual Bengali) instances (Macro-F1 score 0.803).

\begin{table}[ht]
\centering
\scriptsize
\begin{tabular}{|c|c|c|c|c|c|}
\hline
\textbf{Model}              & \textbf{Acc} & \textbf{M-F1} & \textbf{F1 (H)} & \textbf{F1 (O)} & \textbf{AUROC} \\ \hline
\textbf{m-BERT}     &\underline{0.777}  & \underline{0.754}  & \textbf{0.775}  & \underline{0.647}  &	\textbf{0.893}  \\ \hline
\textbf{XLM-Roberta} &\textbf{0.796} & \textbf{0.764} & \underline{0.757} & \textbf{0.649} & \underline{0.891}  \\ \hline
\textbf{MuRIL}     &0.771 &0.722 &0.728 &0.586 &0.830 \\ \hline
\textbf{IndicBERT} &0.723 &0.671 & 0.650 & 0.540 &0.826 \\ \hline
\end{tabular}
\caption{Performance on Translated Data. M: Macro, Acc: Accuracy, H: Hate, O: Offensive.}
\label{bengaliEnglisgTrans}
\end{table}

\begin{table}[ht]
\centering
\scriptsize
\begin{tabular}{|c|c|c|c|c|c|}
\hline
\textbf{Model}              & \textbf{Acc} & \textbf{M-F1} & \textbf{F1(H)} & \textbf{F1(O)} & \textbf{AUROC} \\ \hline
\textbf{m-BERT}     & \textbf{0.856}            &\textbf{0.811}             & \underline{0.694}                    & \textbf{0.852}                 & \textbf{0.930}            \\ \hline
\textbf{XLM-Roberta} & \underline{0.849}             & \underline{0.799}             & \underline{0.670}                    & \underline{0.847}                 & \underline{0.910}            \\ \hline
\textbf{MuRIL}     & 0.845             & 0.791             & 0.647                    & 0.844                 & 0.895            \\ \hline
\textbf{IndicBERT} & 0.839             & 0.787             & 0.649                   & 0.830                 & 0.911            \\ \hline
\end{tabular}
\caption{Performance of Roman Bengali on Joint Training with the Translated Data. M: Macro, Acc: Accuracy, H: Hate, O: Offensive.}
\label{jointTransCodeMixedPerform}
\end{table}

\subsection{Performance of joint training with language transfer}
In this scenario we investigate, even though models trained on translated Bengali instances cannot outperform the monolingual models trained on gold labels, can it be useful to improve the performance of Roman Bengali data? This is motivated by the fact that in a romanized(code-mixed) scenario, people mix English words/phases while writing. Table \ref{jointTransCodeMixedPerform} shows the results on the code-mixed test set. We monitor the performance of m-BERT (Macro-F1 score: earlier (0.790), now (0.811)) and MuRIL (Macro-F1 score: earlier (0.788), now: (0.791)) and observe that these have improved for the detection in the Roman Bengali dataset. However, for XLM-Roberta (Macro-F1 score: earlier (0.805), now (0.799)) and  IndicBERT (Macro-F1 score: earlier (0.793), now (0.787)) the models perform slightly worse compared to those trained on only Roman Bengali gold data. Overall, it can be concluded that while some models are able to leverage the strength of the translated Bengali data while predicting the labels of the Roman Bengali posts, others are not. This might hint at the differences in the generalizability powers of these models. To understand this better, in section \ref{errorAna} we deep dive into the models further using error analysis techniques.

\begin{table}[]
\centering
\resizebox{0.7\linewidth}{!}{%
\begin{tabular}{|l|l|l|l|}
\hline
\textbf{Train} & \textbf{Test} & \textbf{Acc} & \textbf{MF1} \\ \hline
Romin & Romin & 0.905 & 0.894 \\ \hline
Romin & Ours & 0.646 & 0.646 \\ \hline
Ours & Ours & 0.846 & 0.843 \\ \hline
Ours & Romin & 0.774 & 0.754 \\ \hline
\multirow{2}{*}{Joint} & Romin & 0.910 & 0.899 \\ \cline{2-4} 
 & Ours & 0.837 & 0.835 \\ \hline
\end{tabular}%
}
\caption{Comparison with existing dataset~\cite{romim2021hate}. Acc: Accuracy, MF1: Macro-F1}
  \label{tbl:binaryComp}
\end{table}

\section{Additional experiment}
In addition, we perform another experiment to further compare the quality of our dataset with the existing dataset of Romim et al.~\cite{romim2021hate}. Using their dataset, we train the XLM-Roberta model\footnote{We consider XLM-Roberta, as this performs the best while training standalone.} and test its performance on our dataset. Likewise, we test the performance on their dataset when the model is trained on our dataset. We only conduct this experiment with the actual Bengali tweets to have valid comparison with their dataset. We combine hate and offensive into a single class for this experiment, as the authors in~\cite{romim2021hate} have considered these two labels as same.  In Table \ref{tbl:binaryComp} we summarize the results. We observe our model achieves macro-F1 score of 0.754 on their dataset, while the model trained on their dataset achieves 0.646 macro-F1 score on our dataset. Further, we train the XLM-Roberta model jointly with both datasets.  We observe jointly training the model further improved the performance on the Romim et al.~\cite{romim2021hate} test data; however, we do not see any improvement in our test data.

\begin{table*}
\centering
  \includegraphics[width=0.8\linewidth]{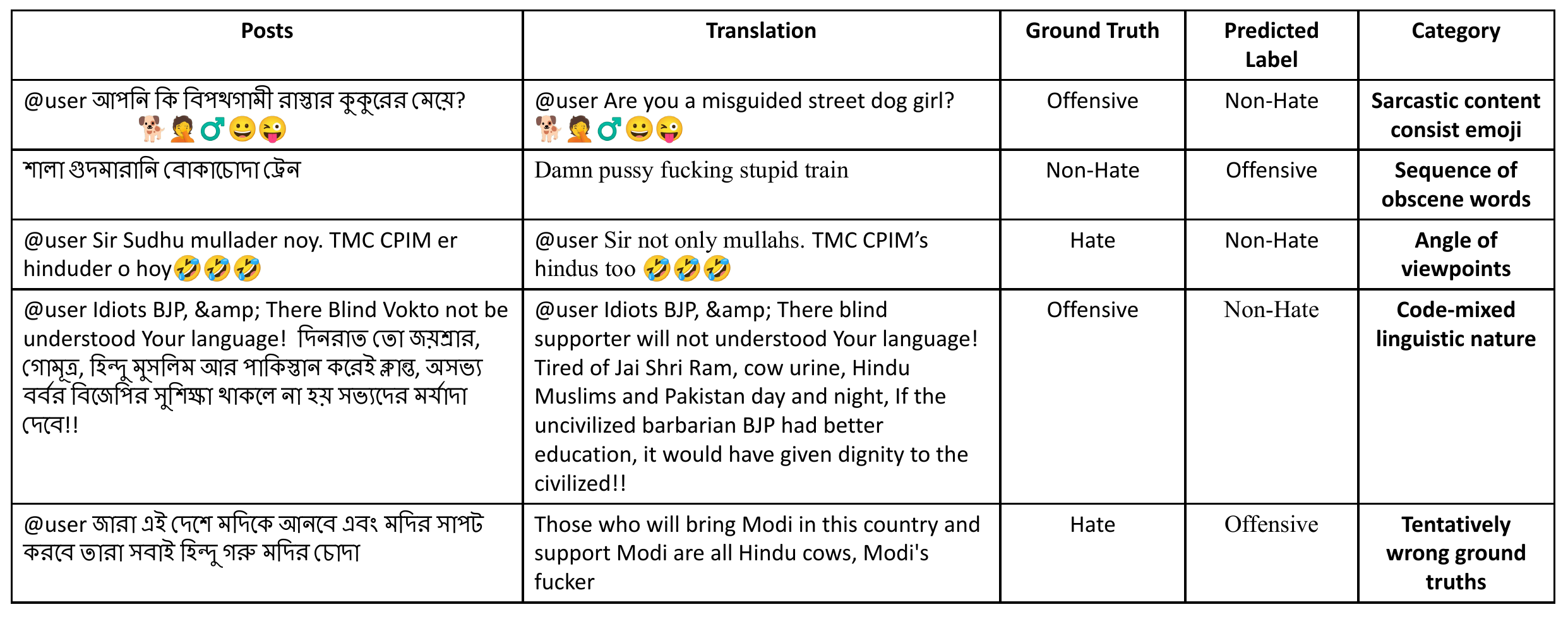}
    \caption{Error analysis on XLM-Roberta (we found similar trend on other models).}
  \label{tbl:manError}
\end{table*}

\section{Error analysis}
\label{errorAna}
In order to deep dive into the models further, we conduct a manual error analysis on our models by using a sample of 50 posts where the model incorrectly categorizes some test instances. We analyze common errors and classify them into the following five categories.

\begin{itemize}

\item \textbf{Sarcastic content consisting emojis}: Communication via emojis is becoming extremely popular these days. Sometimes these emojis completely change the interpretation of the post by making it sarcastic/ambiguous. This naturally results in mis-classification.
\item \textbf{Sequence of obscene words}: Some instances of a series of swear words not targeting individuals or communities are mis-classified. This indicates that the presence of hateful, obscene keywords should not be the only decisive factor for a model to make its predictions. 
\item \textbf{Viewpoints}: Some instances mostly relating to a political or religious sense cannot be fully binary or ternary. The annotators' viewpoint plays a key role in such instances and makes the models to mis-classify these instances. All the models suffer similarly here.

\item \textbf{Code-mixed linguistic structure}: Instances following the grammatical structure of Bengali but written using English words sometimes get mis-classified due to the code-mixed nature of data at hand because there is a heavy between the tokens from Bengali and English.
\item \textbf{Tentatively wrong ground truths}: Some instances containing slur words many not be targeting any group as such. However annotators tentatively marked it hateful leading to the model mis-classifying the post.
\end{itemize}
\par

In Table \ref{tbl:manError} we present example instances for the above categories and the predictions thereof. Though we show the predictions for XLM-Roberta, all the other models also produce similar results.

\section{Discussion}
In this section we discuss the key insights from our results. We observe that depending on the availability of training data points, the performance of the model varies. When we have sufficient number of training instances XLM-Roberta model performs the best. Further we argue that when actual \& Roman Bengali instances are merged together for joint training, models like MuRIL performs the best by leveraging the semantic connection between actual and Romanized instances. This is, to some extent, expected from MuRIL due to the nature of its pre-training mechanism, where both actual language and its transliterated counterpart have been used.

Further exploring the performance of these models in zero-shot setting shows, although XLM-Robera performs best while trained with standalone data, it performs very poorly for unseen data with similar semantic content but a different orthography. In such scenarios, models like m-BERT, IndicBERT exhibit better performance. To improve the performance of these models, when some instances from the target language are used, MuRIL shows an increase in performance at a rate higher than the other models. Observing these results it may be safe to say that when there is a data scarcity for a particular language, it is better to reuse existing fine-tuned models in the same domain. Also careful selection of model is needed. In our case, actual Bengali and Roman Bengali use different characters for writing, but their semantic expressions are same, which is why MuRIL performed best overall.

While doing the in-depth error analysis, we also found that for some cases it can be even difficult for a model to find the actual label correctly. Not only models, as hate speech is complex in nature, sometimes annotators make mistake while labelling them due to differing viewpoints.

\noindent\textbf{Limitation:} There are a few limitations of our work. First is the lack of external context. We have not considered any external context such as profile bio, history of user's posting pattern, gender etc., which might be helpful for the hate speech detection task. Although the effectiveness of these transformer-based models are quite good, they have not been tested against adversarial examples.

\section{Conclusion}

This paper presents a new benchmark dataset for Bengali hate speech detection, consisting of 10K posts from Twitter, covering both actual \& Roman scenarios. Each tweet was annotated with one of the hate/offensive/normal labels. We assessed different transformer-based architectures for hate speech detection. We also experimented with several interlingual transfer mechanisms. Our experiments show how few-shot techniques could be beneficial. Besides, we saw how joint training performs better than training on standalone data. We further notice that joint transliterated training performs best in the case of the Roman Bengali dataset. Our error analysis reveals some of the typical shortcomings of the transformer models. 

As part of the future work, we plan to evaluate the robustness of these models' under adversarial attack as hateful users keep contriving newer ways to deceive the standard hate speech detection models. Another direction could be lessening the biases that can be present in the dataset/model.

\section*{Ethical considerations}
We only analyzed publicly available data crawled via Twitter API. We followed standard ethical guidelines~\cite{rivers-ethical}, not making any attempts to track users across platforms or deanonymize them. We have added a data statement~\cite{bender2018data} in the appendix. Although we achieved good performance and the results look promising, these models cannot be deployed directly on a social media platform without rigorous testing. Further study might be needed to track the presence of unintended bias towards specific target communities.

\bibliography{anthology,custom}

\appendix

\section{Data statement}
\subsection{Curation rationale} 
The dataset consists of a collection of Tweets in actual and roman Bengali. To crawl the dataset, Twitter API has been used.

\subsection{Language variety} The languages of the dataset are in Bengali (bn), Roman Bengali (bn-En).

\subsection{Speaker demographic}
\begin{itemize}
    \item Twitter users
    \item Age: Unknown – mixed.
    \item Gender: Unknown – mixed.
    \item Race/Religion: Unknown – mixed.
    \item Native language: Unknown; Bengali speakers.
    \item Socioeconomic status: Unknown – mixed.
    \item Geographical location: Unknown; mostly from Bangladesh \& India.
\end{itemize}

\subsection{Annotator demographic}
\begin{itemize}
    \item Age: 22-29.
    \item  Gender: 2 male \& 2 female.
    \item Race/Religion: prefer not to disclose.
    \item Native language: Bengali.
    \item Socioeconomic status: undergraduate students.
\end{itemize}

\subsection{Speech situation} Discussions held in public on Twitter platform.

\subsection{Text characteristics} All the sentences in this dataset come from Twitter. 

\subsection{Other} N/A

\end{document}